\documentstyle[12pt,macros,a4wide,fullname]{article}
\input{psfig-scale}
\newcommand{\epsfig}[2]{\begin{figure}[htbp]
\centerline{\psfig{figure=#1.eps,scale=50}}\caption{#2}\label{#1}
\end{figure}}
\let\prg=\sf
\let\em=\bf
\newcommand{\mk}[1]{\mbox{$\overline{\mbox{\prg#1}}$}}

\newcommand{\bul}{{\footnotesize$\bullet$}}
\newcommand{\dt}{$\cdot$}
\begin{document}
\title{\mbox{Emergent Parsing and Generation}
\mbox{with Generalized Chart}\thanks{Presented at
the 15th International Conference
on Computational Linguistics (COLING\,'94) in Kyoto,
August 5th to 9th, 1994.}}
\author{{\bf HASIDA K\^oiti}\\Electrotechnical Laboratory\\
1-1-4 Umezono, Tukuba, Ibaraki 305, Japan\\
%\sl Tel:\,+81-298-58-5928, Fax:\,+81-298-58-5930, 
\sl E-mail:\,hasida@etl.go.jp}
\date{}
\maketitle

\begin{abstract}
A new, flexible inference method for Horn logic program is proposed,
which is a drastic generalization of chart parsing,
partial instantiation of clauses in a program roughly
corresponding to arcs in a chart.
Chart-like parsing and semantic-head-driven generation emerge from this method.
With a parsimonious instantiation scheme for ambiguity packing,
the parsing complexity reduces to that of standard chart-based algorithms.
\end{abstract}

\section{Introduction}

Language use involves very complex interactions among very diverse
types of information, not only syntactic one
but also semantic, pragmatic, and so forth.
It is hence inappropriate
to assume any specific algorithm for syntactic parsing or generation,
which prescribes particular processing directions
(such as left-to-right, top-down and bottom-up)
and is biased for specific types of domain knowledge
(such as a context-free grammar).
To account for the whole language use,
we will have to put many such algorithms together,
ending up with an intractably complicated model.

A better strategy is to postulate
no specific algorithms for parsing or generation or any particular task,
but instead a single uniform computational method from which
emerge various types of computation including parsing and generation
depending upon various computational contexts.

For example, Earley deduction \cite{pereira&warren83} is a general
procedure for dealing with Horn clauses which gives rise to
Earley-like parsing when given a context-free grammar
and a word string as the input.
\namecite{shieber88} has generalized this method so as to adapt to
sentence generation as well.
Those methods fail to give rise to efficient computation
for a wide variety of contexts, however,
because they prescribe processing directions
such as left-to-right for parsing and bottom-up for generation.
They also lack a general way of efficient ambiguity packing
unlimited to context-free grammars.
\namecite{chpgwB} proposes a more general inference method for
clausal form logic programs
which accounts for efficient parsing and generation as emergent phenomena.
This method prescribes no fixed processing directions,
and the way it packs ambiguity is not specific to context-free grammars.
However, it is rather complicated and
has greater computational complexity than standard algorithms do.

In this paper we propose another inference method for
Horn logic programs based on \namecite{chpgwB},
and show that efficient parsing and generation emerge from it.
Like that of \namecite{chpgwB}, this method is totally constraint-based
in the sense that it presupposes no fixed directions of information flow,
but it is more efficient owing to a parsimonious method of instantiation.
In \sect{gchart} we define this inference method,
which is a generalization of chart parsing,
and may be also thought of as a connection method or
a sort of program transformation.
\sect{pg} illustrates how efficient parsing and generation
emerge from this method without any procedural stipulation specific
to the task and the domain knowledge (syntactic constraints).
\sect{pack} introduces
a parsimonious instantiation method for ambiguity packing.
We will show that owing to this method the efficiency
reaches that of the standard algorithms with regard to context-free parsing.
\sect{conc} concludes the paper by touching upon further research directions.

\section{Partial Instantiation\label{gchart}}

A constraint is represented in terms
of a Horn clause program such as below.
\begin{blist}
\item[(a)] {\prg--p(A,B) --A=a(C).}
\item[(b)] {\prg p(X,Y) --X=a(Y).}
\item[(c)] {\prg p(U,W) --p(U,V) --p(V,W).}
\end{blist}
Names beginning with capital letters represent variables,
and the other names predicates and functors.
The atomic formulae following the minus sign are negative (body) literals,
and the others are positive (head) literals.
A clause without a positive literal is called a {\em top clause},
whose negation represents a goal (top-level hypothesis),
which corresponds to a query in Prolog.
For instance, top clause (a) in the above program
is regarded as goal $\exists A,B,C\{p(A,B) \wedge A=a(C)\}$.
In general, there may be several top clauses.
The purpose of computation is to tell whether any goal is satisfiable,
and if so obtain an answer substitution for the terms (variables) in
a satisfiable goal.
We consider the minimal Herbrand models as usual.
So the set of answer substitutions for {\prg A} in the above program is
\{{\prg a(B), a(a(B)), a(a(a(B))), $\cdots$}\}.

A graphical representation of this program is shown in \fig{par0}.
\epsfig{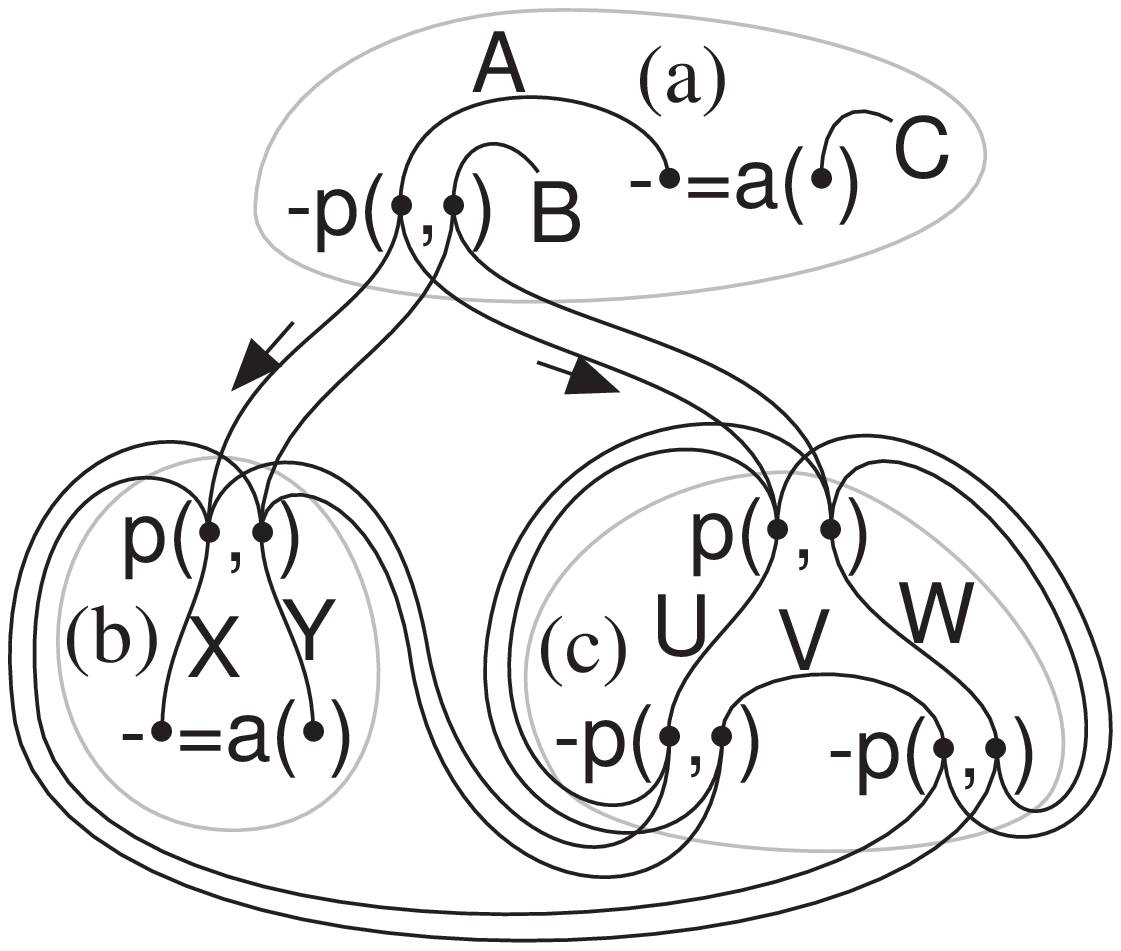}{A graphical representation of a program.}
Here each clause is the set of the literals enclosed in a dim closed curve.
A link connecting arguments in a clause is the term (variable)
filling in those arguments.
(It is a hyperlink when there are more than two arguments.)
A transclausal link represents the unifiability between
two corresponding arguments of two unifiable literals.
(Neglect the arrows for a while.)

A {\em hypothesis} is a conjunction of atomic formulas and bindings.
The premise of a clause (i.e., the conjunction
of the atomic formulas and bindings which appear as negative literals)
is a hypothesis.
An {\em expansion} for a hypothesis is a way of combining
(instances of) clauses by resolutions so as to translate the hypothesis to
another hypothesis involving bindings only.
We will refer to an expansion by the sequence of clauses
in the order of leftmost application of resolution
using their instances.\footnote{
Here we mention the order among the literals in a clause
just for explanatory convenience.
This order is not significant in the computation discussed later.}
In the above program, for example, expansion $\langle$c,\,b,\,b$\rangle$
translates the top-level hypothesis
{\prg s(A,B) \AND\ A=a(C)} to a hypothesis {\prg A=a(C) \AND\ C=a(B)}.
An expansion of a clause is an expansion of its premise.
We will simply say `an expansion'
to mean an expansion of the top-level hypothesis.
A program represents a set of expansions,
and the computation as discussed later is to transform it
so as to figure out correct hypotheses while
discarding the wrong expansions (those entailing wrong hypotheses).

We say that there is a {\em dependency} between two terms
when those terms are unified in some expansion,
and the sequence of terms (including them)
mediating this unification is called
the {\em dependency path} of this dependency.
In \fig{par0}, for instance, the dependency between {\prg A} and
{\prg X} is mediated by dependency path {\prg A\dt X},
{\prg A\dt U\dt X}, {\prg A\dt U\dt U\dt X}, and so on.
There is a dependency between {\prg C} and {\prg B}, among others,
because of the unifiability of the two {\prg--\bul=a(\bul)}s,
though this unifiability is not explicitly shown in \fig{par0}.
We say a dependency between two terms is {\em consistent}
when they are not bound by inconsistent bindings.
All the dependencies in \fig{par0} are consistent.

A {\em solution} of the program is
an expansion in which every dependency is consistent.
So the computation we propose in this paper is to transform the given program
in such a way that every dependency be consistent.
To figure out dependencies,
we use a symbolic operation called {\em subsumption},
and {\em delete} the parts of the program which contributes to
wrong expansions only.
For example, suppose there is an inconsistent dependency
between terms $\alpha$ and $\beta$.
We create an instance $\beta'$ of $\beta$ by subsumption operations
to be discussed shortly, so that every expansion
containing an instance of $\beta'$ contains an instance of a dependency
path between $\alpha$ and $\beta$.
We can then delete the clause containing $\beta'$ and probably
some more parts of the program without affecting the declarative
semantics of the program.
Below we will define a computational procedure in such a way that
the set of the possible expansions eventually
represent the set of all the solutions.

Subsumption operation is to create {\em subsumption relationship}.
We regard each part (clause, atomic formula, term, etc.)
of a program as the set of its instances,
and say that a part $\xi$ of the program {\em subsumes}
another part $\eta$ to mean that we {\sl explicitly know}
that $\xi\supseteq\eta$.
We consider that a link is subsumed by $\delta$ if and only if one of the
terms it links is subsumed by $\delta$.
We say term $\delta$ is an {\em origin} of $\eta$
when $\eta$ is subsumed by $\delta$.
In this paper we consider that every origin is a bound term
(the term filling in the first argument of a binding).
Let us say that two clauses (or two literals) are {\em equivalent}
when they are of the same form and for each pair of corresponding terms
the two terms have the same set of origins.

Subsumption relation restricts the possibility of expansions
so that if term $\eta$ is subsumed by another term $\delta$,
then every expansion containing an instance of $\eta$ must also
contain an instance of $\delta$.
Subsumption relation is useful to encode
structure sharing among expansions.
In subsumption-based approaches,
a term may subsume several non-unifiable terms
and thus the first term is shared among the latters.
However, that is impossible in unification-based approaches,
where different expansions cannot share
the same instance of a term or a clause.

A {\em partially instantiated clause} is a clause some of whose
terms is subsumed by another term in possibly another clause.
For instance,
\onecitm[pic]{\prg
a(\mk{A$_i$},Z) --b(\mk{A$_i$},\mk{A$_j$}) --c(\mk{A$_j$},Z).}
is a partial instantiation of the following clause:
\onecitm{\prg a(X,Z) --b(X,Y) --c(Y,Z).}
\mk{A} represents a term subsumed by term {\prg A}.\footnote{
This notation is problematic because it is unclear
whether two occurrences of \mk{A} in a clause denote the same term.
In this paper they always do.}
Hereafter we say just `clause' to refer to both uninstantiated clauses
and partially instantiated clauses.

A program consisting of such clauses
is a generalization of a chart \cite{kay80}.
A chart is a graph whose nodes denote positions between words in a sentence
and whose arcs are regarded as context-free rules
each instantiated partially with respect to at most two such positions.
For instance, an active arc from node $i$ to node $j$ labelled with
$[A\rightarrow \bullet\,B\,\bullet\,C]$ is
an instance of rule $A\rightarrow B\,C$ with both sides of $B$
instantiated by positions $i$ and $j$.
This arc approximately corresponds to \pref{pic}.\footnote{
However, an arc in a chart does not precisely correspond to
a partially instantiated clause derived from
a program encoding a context-free grammar in a standard way.
See \sect{pack} for further discussion.}

A {\em subsumption operation} is to extend subsumption relation
by possibly creating a partially instantiated clause.
A subsumption operation is characterized by
the {\em origin}, the {\em source}, and the {\em target}.
The origin (let it be $\delta$) is a bound term.
The source ($\sigma$) and the target ($\tau$) are arguments.
$\sigma$ should already be subsumed by the origin, but $\tau$ should not be so.
They should be connected through a transclausal link $\lambda$.
Let the literal containing $\sigma$ be $\rho$.
Also let the literal containing $\tau$ be $\pi$,
and the clause containing them be $\Phi$.
There are two cases for subsumption, and in both cases
$\sigma$ comes to be linked through $\lambda$
with an argument which is an instance of $\tau$ subsumed by $\delta$.

In the first case of subsumption operation, which we call {\em unfolding},
a partial instantiation $\Phi'$ of $\Phi$ is created.
They are equivalent except that the instance $\tau'$ of $\tau$ in $\Phi'$
is subsumed by $\delta$.
After the unfolding, $\sigma$ is linked through $\lambda$ to
the instance of $\tau$ in $\Phi'$ instead of the original $\tau$,
and accordingly $\rho$ is linked to the instance of $\pi$ in $\Phi'$.
Let $\tau''$ be $\tau$ after the unfolding.
Then $\tau'\cup\tau''=\tau$, $\tau'\cap\tau''=\emptyset$,
and $\tau'=\tau\cap\sigma$ hold.
This implies $\tau'\subseteq\delta$ and $\tau''\cap\sigma=\emptyset$.
So $\tau''$ and $\sigma$ are not unifiable.

For instance, the two subsumption operations indicated by the two
arrows in \fig{par0} are unfoldings.
In either case, the origin and the source are both {\prg A}.
The target in the left is {\prg X} and that in the right is {\prg U}.
We obtain the program in \fig{par1}
\epsfig{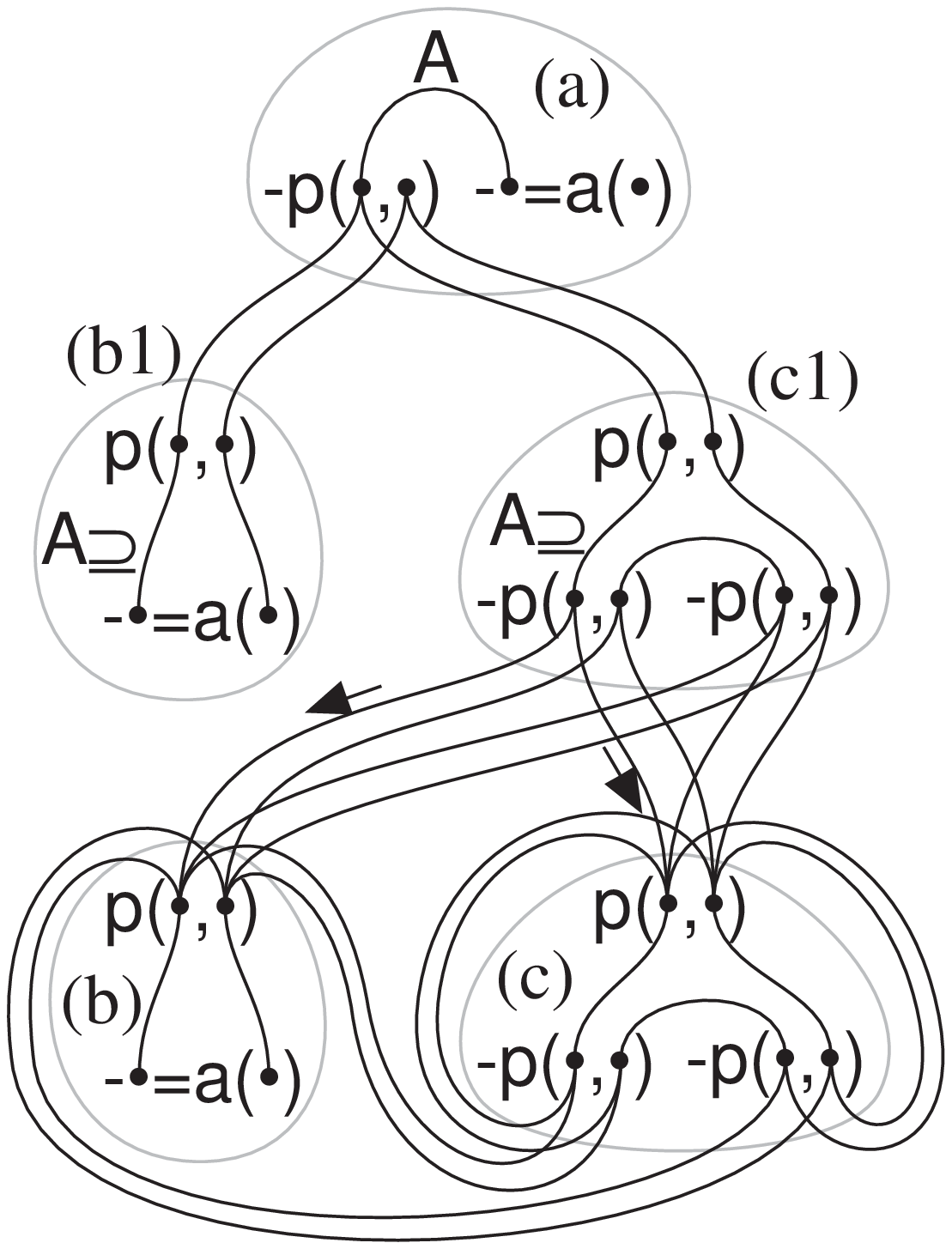}{After subsumptions to {\prg X} and {\prg U} by {\prg A}.}%
by these operations,
where partial instantiation (b1) and (c1) of (b) and (c) have been created,
respectively.

In \fig{par0}, the subsumption operation through the (invisible)
link connecting {\prg C} and {\prg Y} is not executable now,
because the unification represented by this link presupposes
the unification of {\prg A} and {\prg X}
through the dependency paths {\prg A\dt X}, {\prg A\dt U\dt X},
{\prg A\dt U\dt U\dt X}, and so on.
That is, it is only when {\prg C} subsumes an instance (let it
be {\prg Y$'$}) of {\prg Y}
that subsumption from {\prg C} to {\prg Y$'$} is possible.
(This subsumption is an unfolding without any copy,
because then {\prg C} automatically subsumes {\prg Y$'$}.)
Same for the subsumption in the opposite direction.

The second case of subsumption operation is called {\em folding}.
It takes place when there is already a literal $\pi'$ equivalent
to $\pi$ except that its argument $\tau'$ corresponding to $\tau$
is subsumed by $\delta$.
In this case, no new instance of clause is created, but instead link $\lambda$
is switched so that it links $\sigma$ with $\tau'$ and accordingly
$\rho$ is linked with $\pi'$.
Let $\tau''$ be $\tau$ after the unfolding.
Then $\tau\cap\tau'=\emptyset$ both before and after the folding,
and $\sigma\cap\tau$ is subtracted from $\tau$ and added to $\tau'$
by the folding.
Folding is triggered when there exists literal $\pi'$ as described above,
and unfolding is executed otherwise.
If there existed several such $\pi'$s, folding takes place,
creating as many instances of $\lambda$ and connecting to those $\pi'$s.

The two subsumption operations indicated in \fig{par1} are foldings.
Actually, in the left, the {\prg p(\bul,\bul)} in (b1) and that in (b)
are equivalent except that the first argument of the former is
subsumed by {\prg A}.
So the link with the arrow and the parallel accompanying link are
switched up to {\prg p(\bul,\bul)} in (b1).
Similarly for the right subsumption.
Shown in \fig{par2}
\epsfig{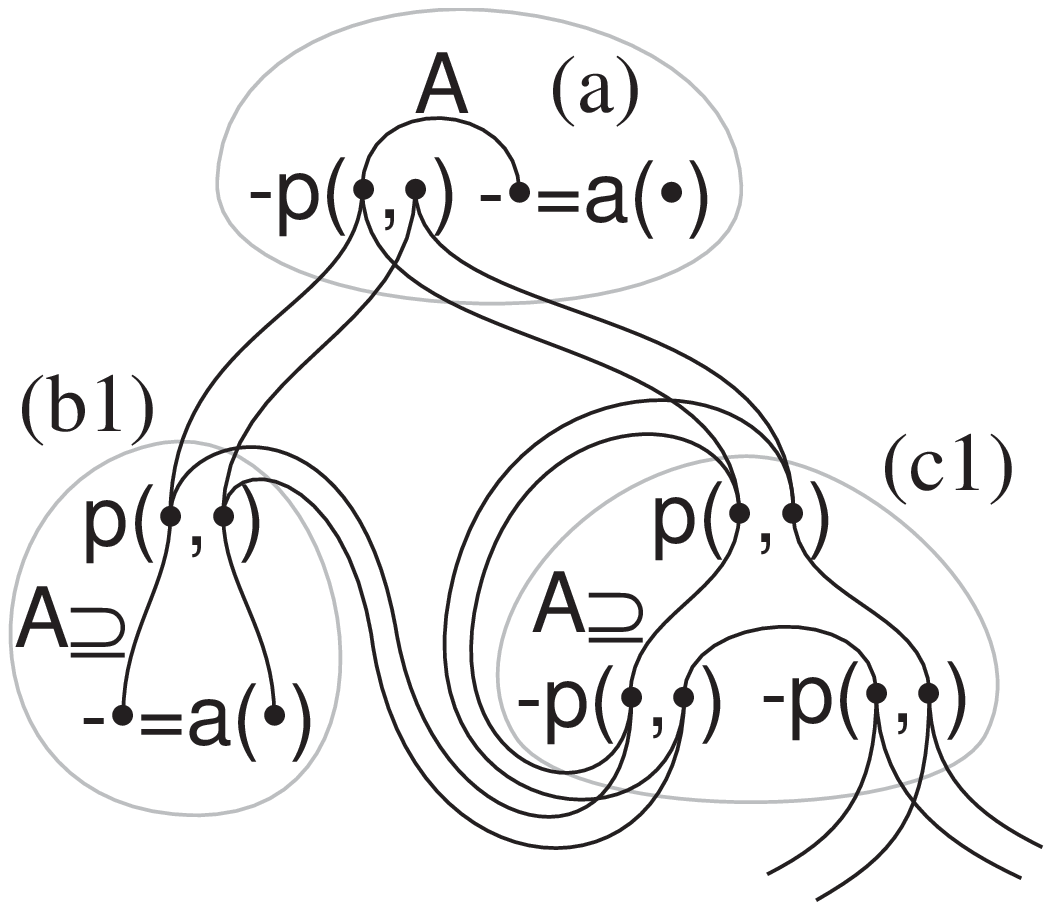}{After foldings.}%
is the result.

Note that the original program encodes a problem of partial parsing
of a string beginning with ``a'' under the context-free grammar consisting of
the following rules.
\begin{quote}$
P\rightarrow {\rm a}\\
P\rightarrow P\,P$
\end{quote}
The result in \fig{par2} encodes the infinitely many possible parses
of this incomplete sentence.
Note also that here the subsumption from {\prg C} to the instance of
{\prg Y} in (b1) would be possible if {\prg C} were bound.
The next section contains relevant examples.

When a link is subsumed by two terms bound by two inconsistent
bindings (such as {\prg \bul=a} and {\prg \bul=b}),
then that link is {\em deleted},
surrounding clauses possibly being deleted if some of their atomic
formulas are linked with no atomic formula any more.

For the sake of simplicity, we mainly consider
{\em input-bound} programs in this paper.
We say a program is input-bound when every dependency
path between bound terms connects a term in a top clause
and one in a non-top clause.
The program in \fig{par0} and the ones for parsing and generation
in the following section are all input-bound programs.
For input-bound programs,
we have only to consider subsumptions by terms in top clauses:
input-driven computation.
Also, in input-driven computation for input-bound programs
we do not have to worry about duplications of origins by subsumptions.

Both subsumption and deletion preserve
the declarative semantics of the program (the set of the solutions),
though we skip a detailed proof due to the space limitation.
So when they are not applicable any more,
every expansion is a solution and vice versa.
For input-bound programs, the input-driven computation always terminates
within time polynomial as to the size of the program.
This is because there are at most $n^m$ partially instantiated
clauses derived from a clause with $m$ terms,
where $n$ is the size of the input
(the number of bound terms in the top clause(s)),
and accordingly there are polynomially many transclausal links.
Obviously, partially instantiated clauses and new transclausal links are
each created in constant time.
It is also clear that each folding terminates in polynomial time.

\section{Parsing and Generation\label{pg}}

Here we show that chart-like parsing and semantic-head-driven
generation emerge from the above computational method.
We discuss examples of parsing and generation
both on the basis of the following grammar.
\begin{blist}
\citm[defs]{\prg s(Sem,X,Z) --np(SbjSem,X,Y) --vp(Sem,SbjSem,Y,Z).}
\citm[defvp]{\prg vp(Sem,SbjSem,X,Z)
	--v(Sem,SbjSem,ObjSem,X,Y) --np(ObjSem,Y,Z).}
\citm[defnp]{\prg np(Sem,X,Y) --Sem=tom --X="Tom"(Y).}
\citm{\prg np(Sem,X,Y) --Sem=mary --X="Mary"(Y).}
\citm{\prg v(Sem,Agt,Pat,X,Y) --Sem=love(Agt,Pat) --X="loves"(Y).}
\end{blist}
Since we have already mentioned ambiguity packing in the previous section,
below we do not explicitly deal with ambiguity but
instead discuss just one sentence structure in both parsing and generation.

Let us first consider parsing of sentence `Tom loves Mary'.
The problem is encoded by the program in \fig{PARSE}.
\epsfig{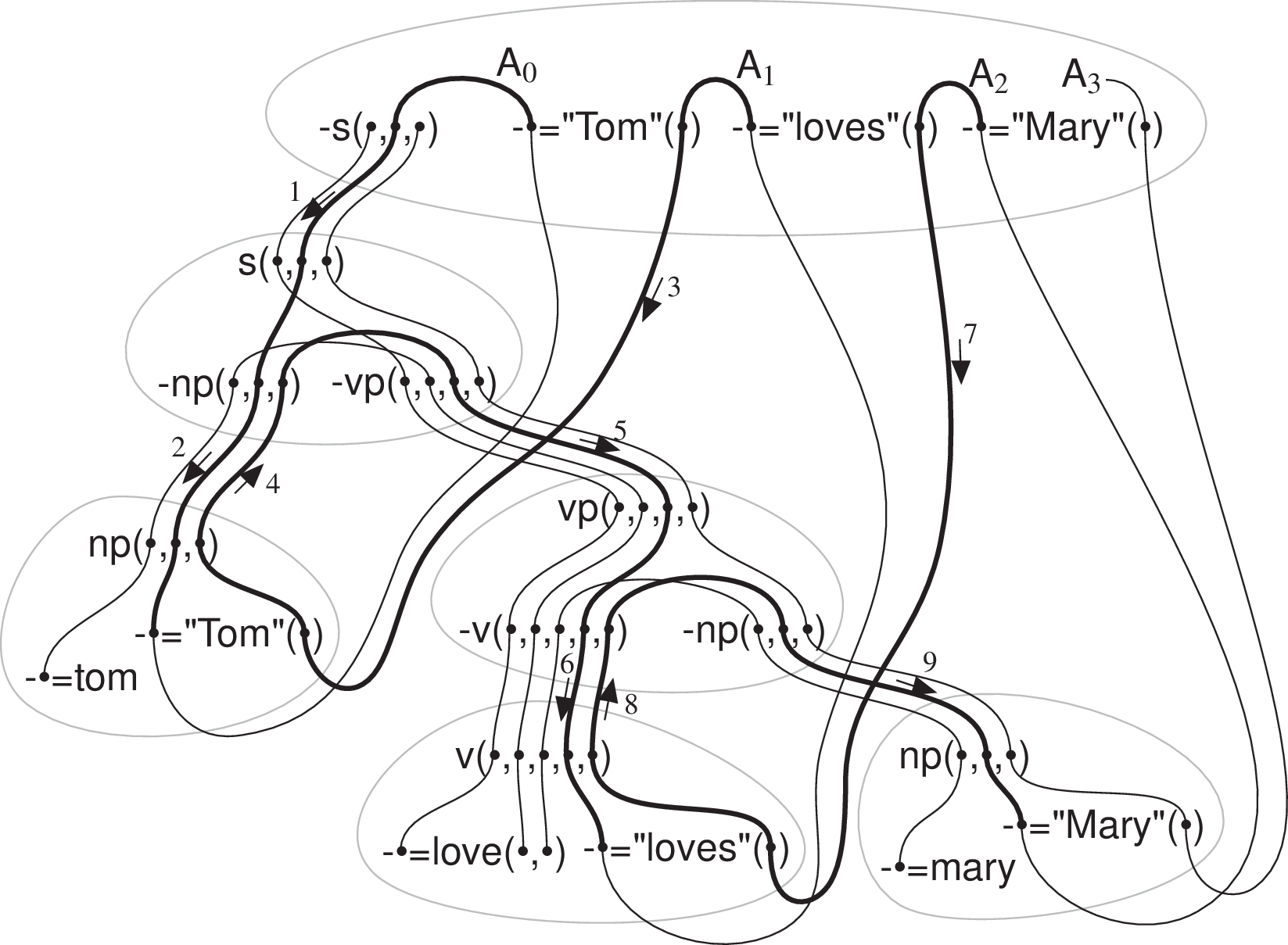}{Parsing}%
The input-driven computation proceeds as shown by the arrows,
which represent subsumption operations
taking place in the ordering indicated by the labelling numbers.
A thick dependency path is processed by
successive subsumptions with the same origin.
The only subsumption operations executable in the initial situation
is the one numbered 1 and after that the one numbered 2,
along the thick path between {\prg A$_0$} and {\prg X} in \pref{defnp}.
As the result of these unfoldings, we obtain the following clauses.
\begin{blist}
\citm[s0]{\prg
s(Sem,\mk{A$_0$},Z) --np(SbjSem,\mk{A$_0$},Y) --vp(Sem,SbjSem,Y,Z).}
\citm[np0]{\prg
\prg np(Sem,\mk{A$_0$},\mk{A$_1$}) --Sem=tom --\mk{A$_0$}="Tom"(\mk{A$_1$}).}
\end{blist}
Of course other partially instantiated clauses may be created here
from definition clauses of {\prg s} other than \pref{defs}
and those of {\prg np} other than \pref{defnp},
but we omit them here and concentrate on just one solution.

Now the copy of link with the arrow numbered 3 connected to \pref{np0} can
mediate subsumption operations.
So the subsumption operation indicated that arrow is triggered,
though that does not duplicate \pref{np0}
because {\prg A$_1$} already subsumes the target.
The result is already reflected in \pref{np0}.
The subsequent subsumption operations numbered 4, 5, and 6 will
yield the following clauses.
\begin{blist}
\citm[s1]{\prg
s(Sem,\mk{A$_0$},Z) --np(SbjSem,\mk{A$_0$},\mk{A$_1$})
	--vp(Sem,SbjSem,\mk{A$_1$},Z).}
\citm[vp0]{\prg
vp(Sem,SbjSem,\mk{A$_1$},Z) --v(Sem,SbjSem,ObjSem,\mk{A$_1$},Y)
--np(ObjSem,Y,Z).}
\citm[v0]{\prg
v(Sem,Agt,Pat,\mk{A$_1$},\mk{A$_2$}) --Sem=love(Agt,Pat)
--\mk{A$_1$}="loves"(\mk{A$_2$}).}
\end{blist}
Now the subsumption operations by {\prg A$_2$} are commenced,
due to the creation of \pref{v0}.
Accordingly, the following clauses are created, and the parsing is finished.
\begin{blist}
\citm{\prg
s(Sem,\mk{A$_0$},\mk{A$_3$}) --np(SbjSem,\mk{A$_0$},\mk{A$_1$})
--vp(Sem,SbjSem,\mk{A$_1$},\mk{A$_3$}).}
\citm{\prg
vp(Sem,SbjSem,\mk{A$_1$},\mk{A$_3$})
--v(Sem,SbjSem,ObjSem,\mk{A$_1$},\mk{A$_2$})
--np(ObjSem,\mk{A$_2$},\mk{A$_3$}).}
\citm{\prg
np(Sem,\mk{A$_2$},\mk{A$_3$}) --Sem=mary --\mk{A$_2$}="Mary"(\mk{A$_3$}).}
\end{blist}

From the earlier discussion, in the case of context-free parsing
the number of clauses created there is $O(n^M)$,
where $n$ is the number of the input words and
$M$ the maximum number of the occurrences of non-terminal symbols
in a context-free rule.
This is larger than the space complexity of the standard parsing algorithms,
but later we will show how to improve the efficiency
so as to be equivalent to the standard algorithms.

No particular order among the subsumption operations is prescribed
in the above computation,
and so it is not inherently limited to top-down or bottom-up.
Note also that the left-to-right processing order among the input words
is derived from the definition strong link,
rather than stipulated as in Earley deduction, among others.
We can account for island-driven parsing as well,
by allowing links between bindings to trigger subsumptions more earlier.

Let us next take a look at sentence generation.
Consider the program shown in \fig{GEN}.
\epsfig{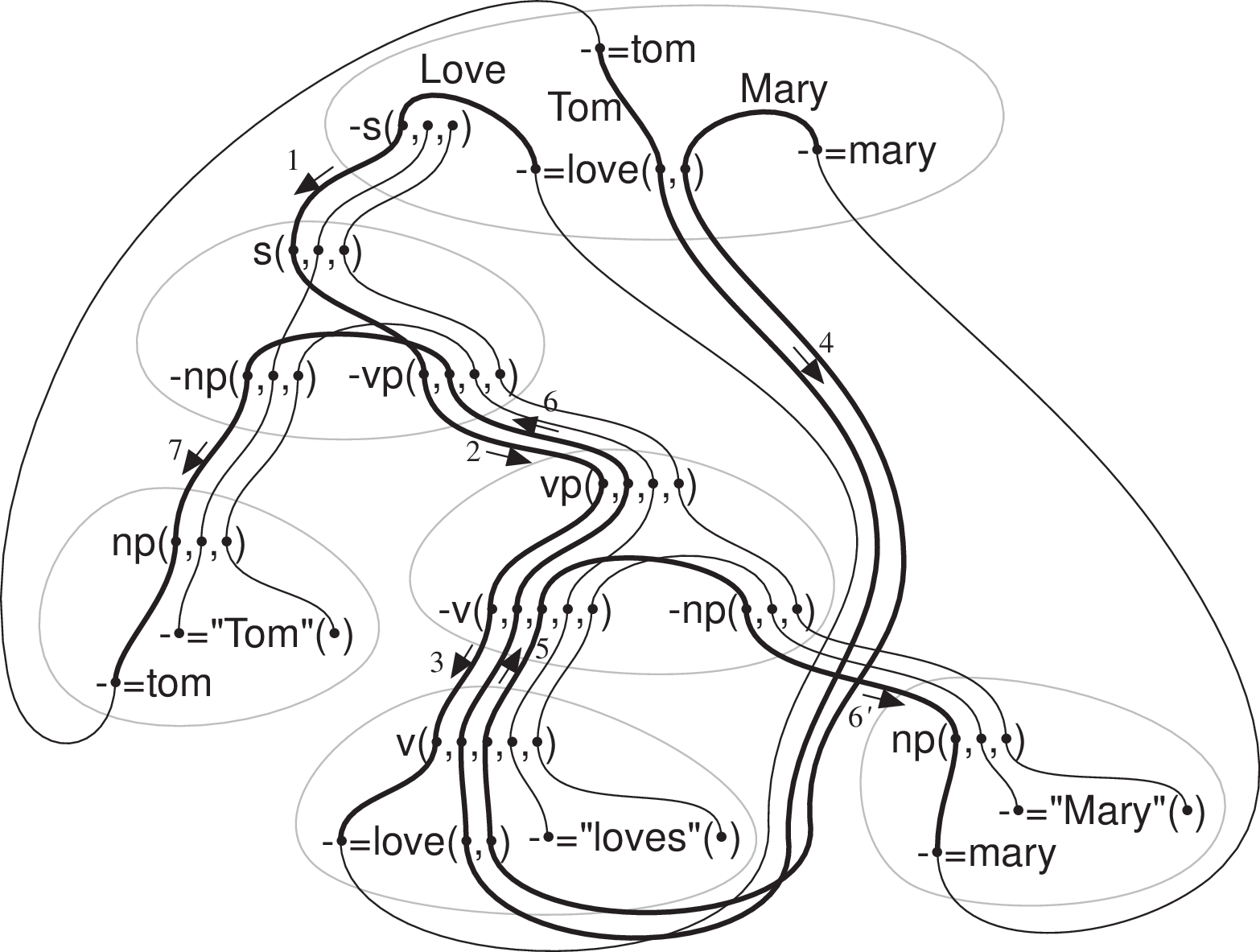}{Generation}%
Here the input is semantic structure {\prg love(tom,mary)}.
Again the computational process is indicated by the numbered arrows.
6$'$ takes place after 5, but the order among 6, 7, and 6$'$ is arbitrary
as long as 6 should be before 7.
So the only possible subsumption operation in the beginning
is the ones by {\prg Love}, which go through
the thick curve connecting {\prg Love} and the {\prg X} in \pref{defvp}.
This creates the following clause, among others.
\onecitm[v1]{\prg 
v(\mk{Love},\mk{Tom},\mk{Mary},X,Y)
--\mk{Love}=love(\mk{Tom},\mk{Mary}) --X="loves"(Y).}
Now subsumption operations can go through
the copies of the other two thick curves.
So we are creating the following clauses, among others.
\begin{blist}
\citm{\prg s(\mk{Love},X,Z) --np(\mk{Tom},X,Y) --vp(\mk{Love},\mk{Tom},Y,Z).}
\citm{\prg vp(\mk{Love},\mk{Tom},X,Z) --v(\mk{Love},\mk{Tom},\mk{Mary},X,Y)
--np(\mk{Mary},Y,Z).}
\citm{\prg np(\mk{Tom},X,Y) --\mk{Tom}=tom --X="Tom"(Y).}
\citm{\prg np(\mk{Mary},X,Y) --\mk{Mary}=mary --X="Mary"(Y).}
\end{blist}

Note that this generation process amounts to a generalization of
semantic-head-driven generation \cite{shieber89}.
The order among the retrievals of semantic heads is
the order of subsumption operations by different terms in
the input semantic structure,
just as with the processing order among words in the case of parsing.%
\footnote{So the semantic-head-driven generation parallels
better with left-to-right parsing than with syntactic-head-driven parsing.}
Also as in the case of parsing,
the computational complexity of such a generation
is polynomial with respect to the size of the input semantic structure,
provided that the program is input-bound and the computation is input-driven.
Although the above example deals with only a single sentence structure,
in general cases ambiguity packing naturally takes place
just as with parsing of ambiguous sentences.

Under the restriction that the program be input-bound,
the grammar cannot employ feature structures prevalent
in the current linguistic theories, and also must be semantically monotonic
\cite{shieber89}\footnote{The semantic monotonicity is practically
same as the input-boundness with regard to semantic structures.}
The proposed method can be generalized so as to remove this restriction,
though the details do not fit in the allowed space.
This generalization makes it possible to deal with feature structures
and semantically non-monotonic grammars.
Of course the computation is not any more generally guaranteed to terminate
(because Horn programs can encode Turing machines),
but our method still has a better termination property
than more simplistic ones such as Prolog interpreter or Earley deduction.
For instance, endless expansion of left recursion or
SUBCAT list, which would happen in simple top-down computations,
is avoided owing to folding.

\section{Incremental Copy\label{pack}}

The parsing process discussed above is computationally
more complex than chart parsing.
Here we improve our method by
introducing a more efficient scheme for ambiguity packing
and thus reduce the parsing complexity to
that of chart parsing, which is $O(n^2)$ for space and $O(n^3)$ for time.

The present inefficiency is due to excessive multiplication of clauses:
much more partially instantiated clauses are created than
arcs in a chart.
So let us suppose that a subsumption operation does not
duplicate a whole clause but only some part of it,
so that a clause is copied incrementally, as shown in \fig{icopy}.
\epsfig{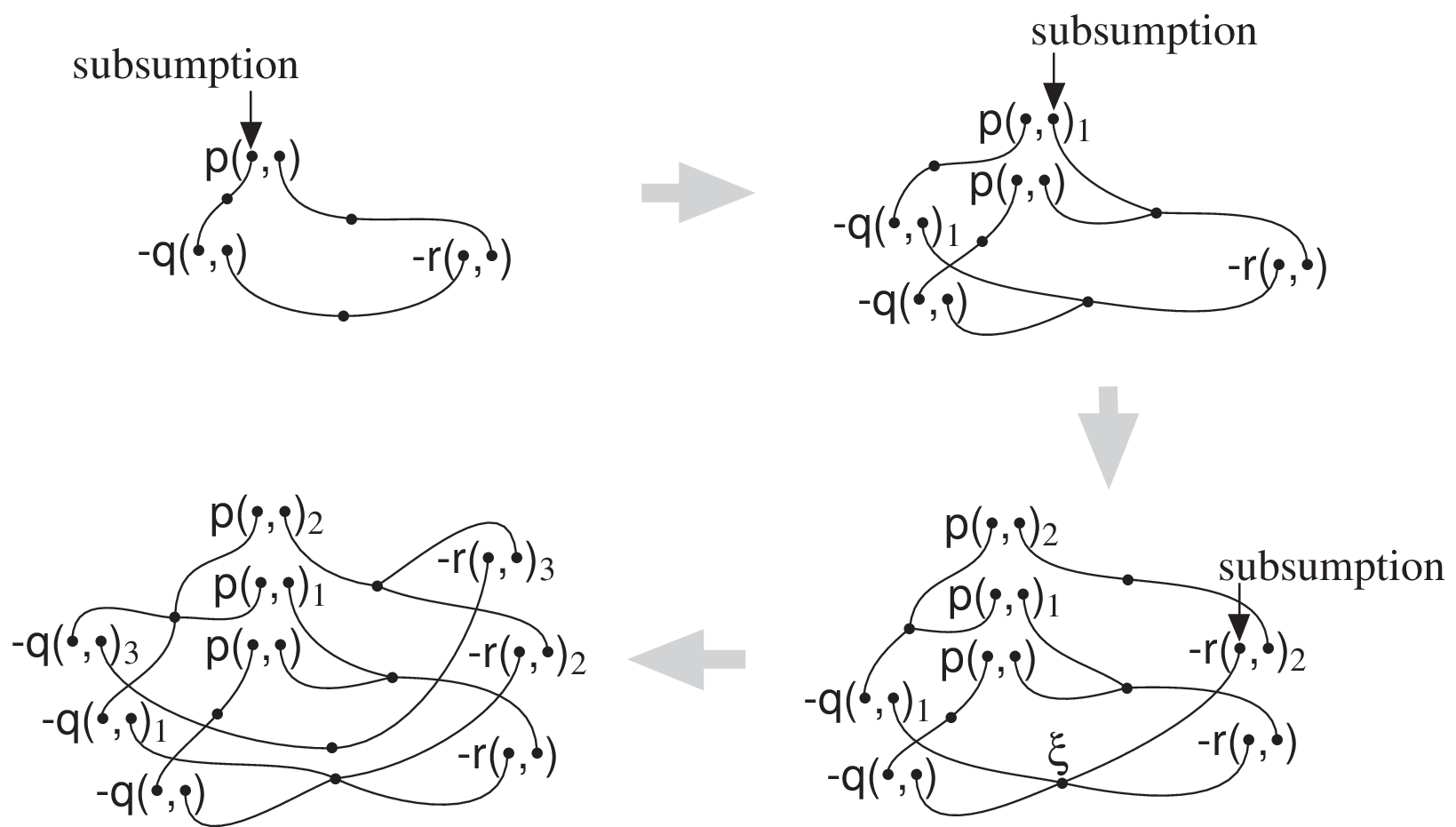}{Subsumptions with Incremental Copy}%
We assume that a subsumption to an argument of a literal copies
the term filling in that argument, the literal,
and some other literals which mention that term,
unless there have already been the terms and literals to be thus created.
Subscript $i$ of a literal indicates that it is created by
the $i$-th subsumption operation.

We must ensure that this partial copying
be semantically equivalent to the copying of whole clauses.
That is a trivial business when
there are just one or two literals in the original clause.
The case where there are more than three literals
reduces to the case where there are exactly three literals,
by grouping several literals connected directly (through terms)
and treat them as if they were one literal.
So below let us consider the case where there are three literals in a clause.

A non-trivial check must be done
in such a case as in the lower right of \fig{icopy}.
Here you must copy {\prg --r(\bul,\bul)$_2$} and {\prg --q(\bul,\bul)$_1$}
but not {\prg --q(\bul,\bul)}, because
{\prg--r(\bul,\bul)$_2$} is {\em compatible} with {\prg --q(\bul,\bul)$_1$}
but not with {\prg--q(\bul,\bul)}.
We say that a set of literals are compatible when
there is an instance of the clause which involves
an instance of each of those literals.
Also, two literals are said to be {\em heterogeneous}
when they have different originals in
the original uninstantiated clause.
(The original of an original literal is itself.)
In general, when a subsumption operation copies two heterogeneous,
directly connected literals and creates two directly connected literals,
the necessary and sufficient condition for this partial copy
to be semantically equivalent to the full-clause copy is obviously
that the former two literals be compatible.

When two of the original literals are not connected directly with each other,
two heterogeneous literals
which have directly connected originals are
compatible iff they are also directly connected;
we need not consider two literals whose originals are not directly connected,
because one subsumption operation does not copy such literals at a time.
When all of the three original literals are
connected directly with each other,
two heterogeneous literals are compatible iff they are connected not only
directly but also through another literal heterogeneous to both.
In fact, {\prg--r(\bul,\bul)$_2$} and {\prg--q(\bul,\bul)$_1$}
are connected both through term $\xi$ and through {\prg p(\bul,\bul)$_2$},
but {\prg--r(\bul,\bul)$_2$} and {\prg--q(\bul,\bul)}
are not connected through any instance of the original {\prg p(\bul,\bul)}.

In the case of context-free parsing, $O(n^2)$ literals are created,
where $n$ is the number of words in the input string,
provided that the origins of subsumptions are
the positions between the input words only,
due to the input-driven computation.
Since there are just a constant times more links than literals,
the space complexity of context-free parsing hence becomes
$O(n^2)$ in our method.
The time complexity is $O(n^3)$, because there are $O(n)$
different ways of making each literal.
Now the correspondence with chart parsing is more exact.
An arc in the chart corresponds to an instantiated literal.
For instance, arc $[A\rightarrow \bullet\,B\,\bullet\,C]$
from node $i$ to node $j$ corresponds to instantiated literal
{\prg--b(\mk{A$_i$},\mk{A$_j$})},
and $[A\rightarrow \bullet\,B\,C\,\bullet]$
from node $i$ to node $j$ corresponds to {\prg a(\mk{A$_i$},\mk{A$_j$})}.
For a context-free rule with more than two symbols
in the right-hand side,
we can group several literals to one as mentioned above
and reduce it to a rule with just two symbols in the right-hand side.

\section{Concluding Remarks\label{conc}}

We have proposed a flexible inference method for Horn logic programs.
The computation based on it is a sort of program transformation,
and chart parsing and semantic-head-driven
generation are epiphenomena emergent thereof.
The proposed method has nothing specific to parsing, generation,
context-free grammar, or the like.
This indicates that there is no need for any special algorithms
of parsing or generation,
or perhaps any other aspect of natural language processing.

The idea reported above has already been partially implemented
and applied to spoken language understanding \cite{sl},
and an account of how the roles of speaker and hearer may switch
in the midst of a sentence \cite{ju}.
Although this line of work has incorporated a notion
of dynamics \cite{dynamicsJ} as the declarative semantics
to control context-sensitive computation,
we are planning to replace dynamics with probability.
For input-bound programs together with input-driven computation,
it is quite straightforward to define probabilistic semantics
as a natural extension of stochastic context-free grammars, among others,
because all the body literals are probabilistically independent
in that case.
We would like to report soon on a general treatment of probabilistically
dependent literals while preserving the efficient structure sharing,
which will guarantee efficient computation and learning.

\let\em=\it
\bibliographystyle{fullname}
\bibliography{93,94}
\end{document}